\RequirePackage{silence}
\documentclass[pdflatex,sn-vancouver-num,iicol]{sn-jnl}

\usepackage[utf8]{inputenc}
\usepackage[T1]{fontenc}
\usepackage{graphicx}%
\usepackage{multirow}%
\usepackage{amsmath,amssymb,amsfonts}%
\usepackage{amsthm}%
\usepackage{mathrsfs}%
\usepackage[title]{appendix}%
\usepackage[dvipsnames]{xcolor}%
\usepackage{textcomp}%
\usepackage{manyfoot}%
\usepackage{booktabs}%
\usepackage{algorithm}%
\usepackage{algorithmicx}%
\usepackage{algpseudocode}%
\usepackage{listings}%
\usepackage{xspace}%
\usepackage{subcaption}

\usepackage{fancyhdr}

\fancypagestyle{firstpage}{
  \fancyhf{} 

  \fancyfoot[C]{%
    \parbox{\textwidth}{\small
      \vspace{20pt} 
       \begin{center}\thepage\end{center}
       \vspace{20pt}
      This preprint has not undergone peer review or any post-submission improvements or corrections. 
      
      The Version of Record of this article is published in `KI - Künstliche Intelligenz`, 
      and is available online at 
      
      \url{https://doi.org/10.1007/s13218-026-00902-6}

    }%
  }
}

\usepackage{xspace}
\newcommand{\eg}{e.\,g.,\xspace}

\usepackage{paralist}

\usepackage[acronym]{glossaries}
\glsdisablehyper   
\newacronym{3dssg}{3DSSG}{3D Semantic Scene Graph}
\newacronym{clip}{CLIP}{Contrastive Language-Image Pretraining}
\newacronym{gnn}{GNN}{Graph Neural Network}
\newacronym{xai}{XAI}{eXplainable AI}
\newacronym{vlfm}{VLFM}{Vision-Language Foundation Model}
\newacronym{vlm}{VLM}{Vision-Language Model}

\newacronym{lierex}{LIEREx}{Language-Image Embeddings for Robotic Exploration}
\newacronym{expris}{ExPrIS}{Knowledge-Level Expectations as Priors for Object Interpretation from Sensor Data}

\begin{document}
\title[LIEREx project report]{\mbox{LIEREx: Language-Image Embeddings for Robotic Exploration}}

\author*[1]{\fnm{Felix} \sur{Igelbrink}
}\email{felix.igelbrink@dfki.de}
\equalcont{These authors contributed equally to this work.}

\author[2,1]{\fnm{Lennart} \sur{Niecksch}
}\email{lennart.niecksch@dfki.de}
\equalcont{These authors contributed equally to this work.}

\author[2,1]{\fnm{Marian} \sur{Renz}}\email{marian.renz@dfki.de}

\author[1]{\fnm{Martin} \sur{Günther}}\email{martin.guenther@dfki.de}

\author[2,1]{\fnm{Martin} \sur{Atzmueller}}\email{martin.atzmueller@uos.de}


\affil[1]{\orgname{German Research Center for Artificial Intelligence (DFKI)}, \orgdiv{Research Department Cooperative and Autonomous Systems (CAS)}, \orgaddress{\city{Osnabrück}, \country{Germany}}}
\affil[2]{\orgname{Osnabrück University}, \orgdiv{Semantic Information Systems Group}, \orgaddress{\city{Osnabrück}, \country{Germany}}}


\abstract{Semantic maps allow a robot to reason about its surroundings to fulfill tasks such as navigating known environments, finding specific objects, and exploring unmapped areas.
Traditional mapping approaches provide accurate geometric representations but are often constrained by pre-designed symbolic vocabularies. The reliance on fixed object classes makes it impractical to handle out-of-distribution knowledge not defined at design time.
Recent advances in \acrlongpl{vlfm}, such as \acrshort{clip}, enable open-set mapping, where objects are encoded as high-dimensional embeddings rather than fixed labels.
In \acrshort{lierex}, we integrate these \acrshortpl{vlfm} with established \acrlongpl{3dssg} to enable target-directed exploration by an autonomous agent in partially unknown environments.}


\thispagestyle{firstpage} 

\maketitle


\section{Introduction}
\label{sec:introduction}

Autonomous mobile robotic agents operating within partially or entirely unknown environments require a high degree of scene understanding to ensure effective and safe operation.
This fundamental capability relies primarily on the construction of a semantic map from sensor data, combining information about the geometry of the environment with details about the objects contained therein, such as their classes and properties~\cite{nuchterSemanticMapsMobile2008b}.
Traditional semantic mapping systems predominantly rely upon rigid, closed symbolic vocabularies, typically defined by a fixed set of object classes.
This limitation restricts the system's flexibility and scalability, making it impractical to handle generic knowledge or concepts not explicitly predefined at design time.

A significant methodological advancement has emerged through the development of \glspl{vlfm}, enabling the effective combination of open vocabularies with multimodal visual data.
The \gls{clip} model~\cite{radford2021learning} is a prominent example of such \glspl{vlfm}.
These models are trained on vast datasets of image-text pairs, allowing them to learn rich multimodal features and map visual concepts and natural language into a joint feature space.
Consequently, this capability enables open-set semantic mapping, where objects are represented not by fixed labels, but by high-dimensional feature embeddings.

The integration of these \glspl{vlfm} into semantic mapping architectures introduces sophisticated querying capabilities that significantly surpass the limitations of conventional, vocabulary-constrained methods.
This fosters novel applications, particularly in target-directed exploration and persistent surveillance within dynamically changing or partially explored environments.
Utilizing open-set semantic queries, robotic agents gain the capacity to interpret and search for arbitrary objects and abstract concepts using natural language.

Within the \acrshort{lierex}\footnote{\url{https://www.dfki.de/en/web/research/projects-and-publications/project/lierex}} project~(\acrlong{lierex}), we are investigating the integration of these \gls{vlfm} advancements with hybrid machine learning methods to advance semantic mapping and autonomous exploration in mobile robotics.
\Acrshort{lierex} leverages the advantages of open-set map representations alongside the existing spatial reasoning capabilities of popular \gls{3dssg}-based semantics.
Specifically, the project focuses on dynamically generating exploration strategies and deriving optimal observation poses using a neural network-based estimation system.
This approach facilitates the efficient verification of search and exploration results, allowing the system to move beyond relying solely on the language query itself.

This endeavor is closely coupled with the \acrshort{expris} project (\acrlong{expris}), sharing infrastructure and the critical adoption of the \gls{3dssg} as the foundational environment representation.

The remainder of this report is organized as follows: Section~\ref{sec:open_set_mapping} presents the open-set mapping approach as the foundation of the architecture, followed by the exploration planning system in Section~\ref{sec:exploration}.
Section~\ref{sec:demonstrator} describes the robotic demonstrator. Finally, we conclude with a discussion of key technical insights in Section~\ref{sec:discussion}.

\section{Open Set Semantic Mapping}
\label{sec:open_set_mapping}

\begin{figure*}
    \centering
    \includegraphics[width=1.0\linewidth]{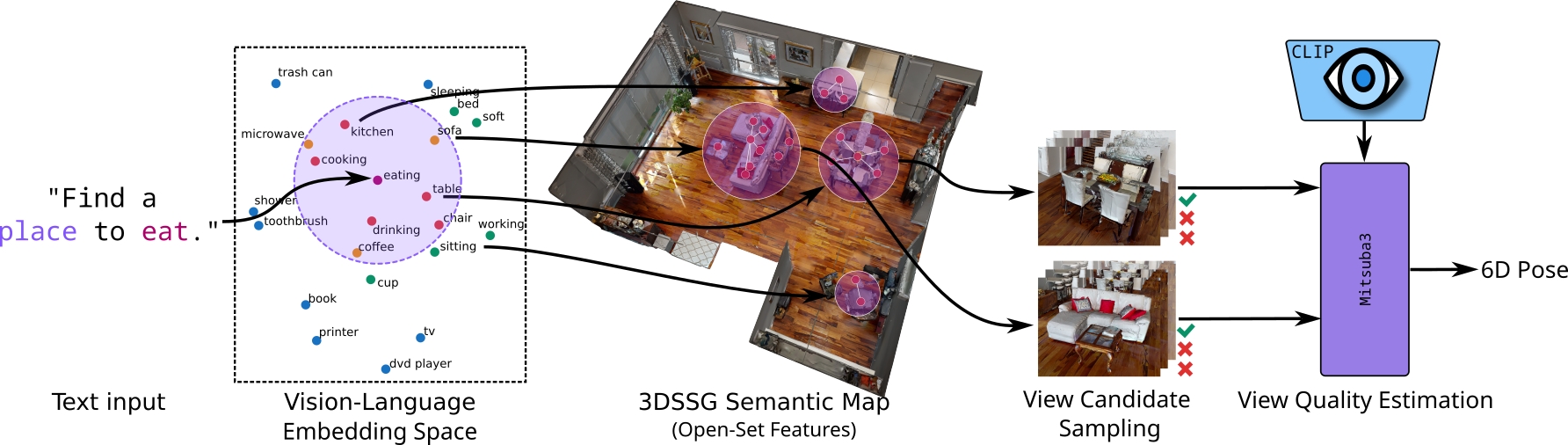}
    \caption{Overview of the \acrshort{lierex} pipeline.
             A textual query retrieves the best matching instances from the \gls{3dssg}, taking into account both the scene context provided by the graph and proximity in the \gls{vlm} embedding space.
             Candidate view poses are then sampled from the \gls{3dssg} and scored by the estimation module based on their semantic similarity to the query.}
    \label{fig:framework}
\end{figure*}

Both \acrshort{lierex} and \acrshort{expris} contribute towards a common, unified semantic framework.
While \acrshort{expris} focuses on deriving a structured representation for integration with existing structured knowledge, \acrshort{lierex} extends and generalizes this goal by incorporating modern \glspl{vlm}.
This allows the system to exploit common sense knowledge already embedded within language concepts in a semantic map.
Specifically, \acrshort{lierex} aims to enhance object retrieval capabilities by not only providing the location or geometry corresponding to a query but also directly generating suitable observation poses where the queried object is likely to be encountered.
These poses can then be utilized by a robotic agent to search for the requested object or location.
The basic pipeline of \acrshort{lierex} is shown in Fig.~\ref{fig:framework}.
Both projects share the \gls{3dssg} as their foundational representation and are designed to be interoperable, combining symbolic knowledge (\acrshort{expris}) with multimodal open-set perception (\acrshort{lierex}).

The structure of our \gls{3dssg} representation is inspired by other popular approaches in the robotics field~\cite{rosinol20203d,hughes2022hydra}.
It is implemented as a heterogeneous graph organized as a dynamic hierarchy of multiple layers.
These layers represent different levels of semantic concepts, ranging from low-level concepts (\eg individual objects) to higher-level concepts (\eg rooms).
A core contribution of \acrshort{lierex} is the extension of this \gls{3dssg} structure to incorporate vision-language features inferred from a \gls{vlfm}.

These models, most notably the pioneering \gls{clip} model~\cite{radford2021learning}, learn rich multimodal feature spaces combining natural language with visual concepts.
The initial \gls{clip} model already showed remarkable performance in zero-shot image classification tasks, even matching supervised models trained on benchmark datasets~\cite{liu2024grounding,ren2024grounding}.
This, along with the model's comparative simplicity and straightforward integration into downstream tasks, has led to the growing popularity of \glspl{vlfm} for many computer vision tasks.
Recently, \textit{Large Language Models} (LLMs) have also emerged as a promising complementary tool in these applications~\cite{zeng2023large}.
Empowered by the integration of \glspl{vlfm}, the new family of \textit{Large Multimodal Models} (LMMs) also allows visual input from images or videos to be processed directly~\cite{huang2024large}.

However, \gls{clip} and similar models, \eg SigLIP~\cite{zhai2023sigmoid}, BLIP~\cite{li2022blip}, and BLIPv2~\cite{li2023blip} only produce feature vectors for the entire image.
This prevents their direct applicability in semantic and instance segmentation tasks required for semantic mapping, as these require fine-grained features to localize detected objects within an image.
Pixel- and region-based approaches~\cite{zhou2022extract,ghiasi2022scaling,ding2022open,luddecke2022image} address this shortcoming by implementing 2D open-vocabulary semantic segmentation for each pixel or larger regions.
Using baseline \glspl{vlfm} and recent instance-agnostic segmentation models~\cite{kirillov2023segment} as foundations, these approaches learn to generate individual feature vectors for pixels or regions, albeit at the cost of significantly higher run times compared to image-based \glspl{vlfm} and traditional models~\cite{yamazaki2024open}.

Unlike traditional segmentation models, these methods do not generate a segmentation mask associated with a specific label unless requested at run-time. Instead, they produce high-dimensional feature maps that encode semantic information directly into the visual representation. This enables a flexible mapping process where visual features are preserved in a latent space and can be retrieved flexibly.

\begin{figure*}[t]
    \centering
    \begin{subfigure}[c]{0.325\linewidth}
        \includegraphics[width=\linewidth]{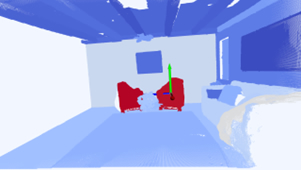}
        \subcaption{}
        \label{subfig:query_1}
    \end{subfigure}
    \begin{subfigure}[c]{0.325\linewidth}
        \includegraphics[width=\linewidth]{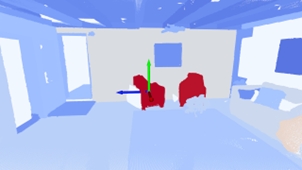}
        \subcaption{}
        \label{subfig:query_2}
    \end{subfigure}
    \begin{subfigure}[c]{0.325\linewidth}
        \includegraphics[width=\linewidth]{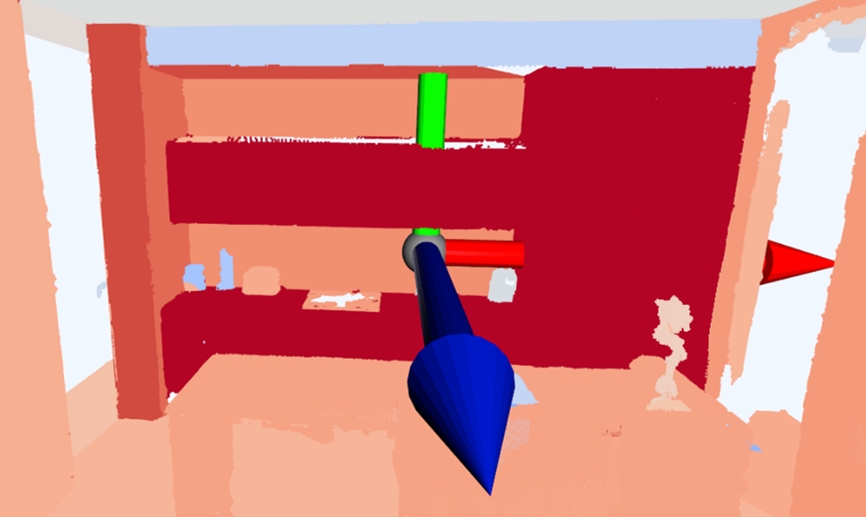}
        \subcaption{}
        \label{subfig:query_3}
    \end{subfigure}
    \caption{Example queries in the VL map.
             (\subref{subfig:query_1}) and (\subref{subfig:query_2}) show the top 2 results for the text query \texttt{chair}, including cropped regions of the surroundings.
             (\subref{subfig:query_3}) depicts a query for the higher-order concept \texttt{kitchenette}, which comprises multiple object instances.}
    \label{fig:queries}
\end{figure*}

In \acrshort{lierex}, we extend our hybrid \gls{3dssg} structure with additional \gls{clip} feature vectors to allow for open-set semantic queries and language-driven reasoning atop the spatial reasoning capabilities provided by the graph.
Rather than relying on distilled models that project \gls{clip} features onto dense pixels or 3D points, we infer feature vectors using a two-step approach aligned with other popular methods~\cite{maggio2024clio,laina2025findanything,kassab2024bare,linok2025beyond}. 
First, potential object candidates are segmented from the incoming RGB-D frames using a class-agnostic segmentation model~\cite{kirillov2023segment, zhao2023fast, zhang2023faster}.
Individual \acrshort{clip} feature vectors are then inferred using the obtained masks and systematically integrated into the corresponding \gls{3dssg} nodes, preferring views where the object of interest is clearly visible~\cite{kassab2024bare}.
The final \gls{3dssg} can be queried using arbitrary text by comparing the query feature vector with the nodes using cosine similarity and returning all matching objects (see Figure~\ref{fig:queries}). 

Additionally, we integrate the spatial structure of the environment with these open-set semantic capabilities. Unlike existing methods that often treat scene graphs as flat collections of objects or a strict hierarchy, our approach aims at inferring the inherent hierarchy of indoor environments automatically. 
By aligning the multi-scale feature maps from the vision-language model with the \gls{3dssg} structure~\cite{maggio2024clio,linok2025beyond}, we ensure that semantic embeddings are not only grounded in visual appearance but also in their spatial context and low-level objects can be clustered dynamically into higher-level objects.
This hierarchical architecture transcends static categorization, facilitating complex reasoning for abstract queries that typically exceed the capacity of standard \glspl{vlfm}.

\section{Exploration Planning System}
\label{sec:exploration}

In the domain of open-vocabulary exploration, recent approaches predominantly focus on augmenting classical frontier-based strategies with semantic guidance derived from \glspl{vlfm} or \glspl{vlm}.
These methods typically maintain the fundamental logic of exploring boundaries between known and unknown space but weight these frontiers based on semantic relevance.

In CoW~\cite{gadre2023cows}, the authors enhance standard frontier selection by integrating \gls{clip}-based relevancy maps to prioritize regions of interest.
Similarly, GOAT~\cite{chang2024goat} employs an image-based object instance memory coupled with a global 2D semantic map to navigate towards target object locations.
To further guide this process, approaches such as~\citet{ren2024explore} project current RGB frames into 3D voxel maps, utilizing \glspl{vlm} to score potential exploration directions based on the identified free space.

A significant limitation common to the majority of these works is their reliance on 2D map representations or 2D projections for planning.
A notable exception is the work by~\citet{laina2025findanything}, which operates directly in 3D space by leveraging \gls{clip}-based cosine similarity to modulate sampling near frontiers.
However, to estimate the information gain of candidate poses, this method relies on ray-casting over TSDF submaps---a process that is computationally expensive and difficult to scale on resource-constrained platforms.

In \acrshort{lierex}, we address these limitations by moving away from purely geometric evaluations of candidate views.
Instead of performing costly online ray-casting for every potential observation pose, we propose a \emph{learned view quality estimation} system.
By training a model to predict the quality of a view relative to a semantic query---based on partially mapped objects and their context---we can efficiently 
predict quality scores for candidate poses directly from the \gls{3dssg} structure.
This allows the agent to verify search results and explore likely object locations with significantly reduced computational overhead.
We integrate this targeted view estimation with a semantically guided frontier-based exploration strategy: the system prioritizes learned view proposals for resolving known or hypothesized semantic targets, while reverting to weighted frontier exploration to uncover the unknown parts of the environment.

\subsection{View Quality Estimation}
\label{sec:view_est}
\begin{figure*}
    \centering
    \includegraphics[width=1.0\linewidth]{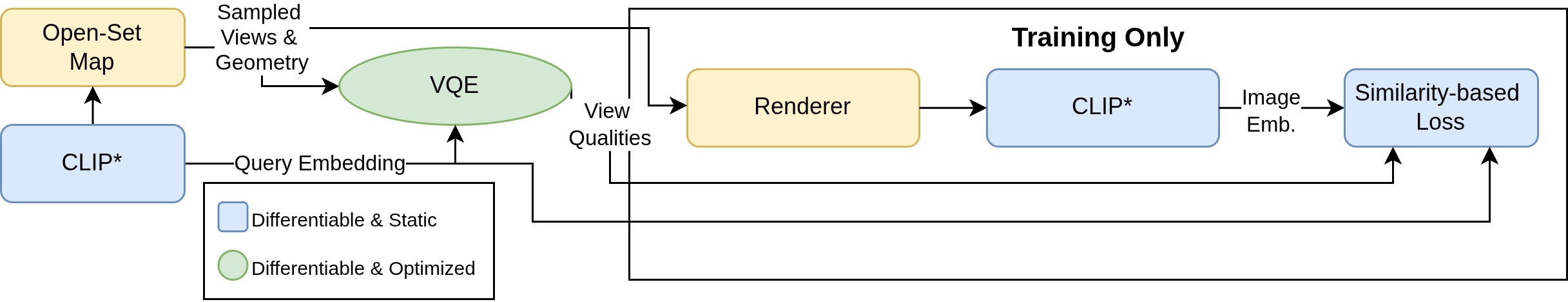}
    \caption{The self-supervised training pipeline for the View Quality Estimation (VQE) module.
             The model learns to predict quality scores by comparing CLIP embeddings of rendered views against the query embedding.}
    \label{fig:training}
\end{figure*}
To enable object-goal navigation, observation poses and trajectories have to be provided in the map's coordinate system.
Traditionally, this has been done using heuristic sampling (\eg FLAP for CAOS~\cite{gedicke2016flap}) or handcrafted cost functions (\eg OK-Robot~\cite{liu2024okrobot}).

The high computational costs for ray-casting and rendering of all poses (compare, \eg~\citet{linok2025beyond}) often makes this infeasible on resource-constrained systems.
A data-driven approach can be much more efficient, but requires appropriate training data.

To the best of our knowledge, there is no public dataset available for the task of view pose estimation in cluttered scenes.
One insight is that a ground truth \emph{best} observation pose is often very hard to define.

In \acrshort{lierex}, we propose a semi-supervised approach based on the evaluation of view quality instead.
Inspired by previous works in grasping~\cite{wang2021graspness, fang2023anygrasp}, we propose an approach that estimates the view quality of uniformly sampled poses around potential targets rather than directly regressing poses.
Instead of focusing on the visibility of single instances, poses are rendered during training and the inferred \gls{clip} embedding of the rendering is compared to the embedding of the query (see Fig.~\ref{fig:training}).
The core idea is that the model should learn to prefer observation poses that clearly distinguish the queried object or concept in the \gls{clip} feature space.

\subsection{Data Generation}
\label{subsec:data_generation}

To facilitate the necessary large-scale training and evaluation of the view quality estimation models, we employ the Habitat simulator~\cite{savva2019habitat}.
In combination with the Matterport3D~\cite{chang2017matterport3d} and Matterport-Habitat (HM3D)~\cite{ramakrishnan2021habitat} datasets this provides a large and diverse set of realistic indoor environments created from high-resolution 3D photogrammetry.

Unlike traditional robotic simulation environments such as Gazebo, Habitat is optimized for efficient testing of perception and planning algorithms without the overhead of simulating full robot dynamics.
This setup enables rapid iteration and large-scale data generation for the self-supervised training of our view pose models.
We utilize the HM3D dataset and generate semantic maps based on ground truth trajectories and semantic annotations using our vision-language mapping pipeline.

\subsection{Training}

During training, query-map pairs are randomly sampled from the set of maps and an expected vocabulary, and the top-k regions are extracted based on cosine similarity of the \gls{clip} embeddings (see Fig.~\ref{fig:queries}).
The views at the evaluated poses are then rendered using the ground truth meshes from the dataset.
\Gls{clip} features of the renderings are computed using a frozen version of the same \gls{clip} model used in the VL mapping pipeline.
Finally, a variant of a cosine loss between the renderings and the query is computed and back-propagated to the model.
Fig.~\ref{fig:training} provides a high-level overview of the proposed data flow.

\section{Indoor Demonstrator}
\label{sec:demonstrator}
\begin{figure}[t]
    \centering
    \includegraphics[width=\linewidth]{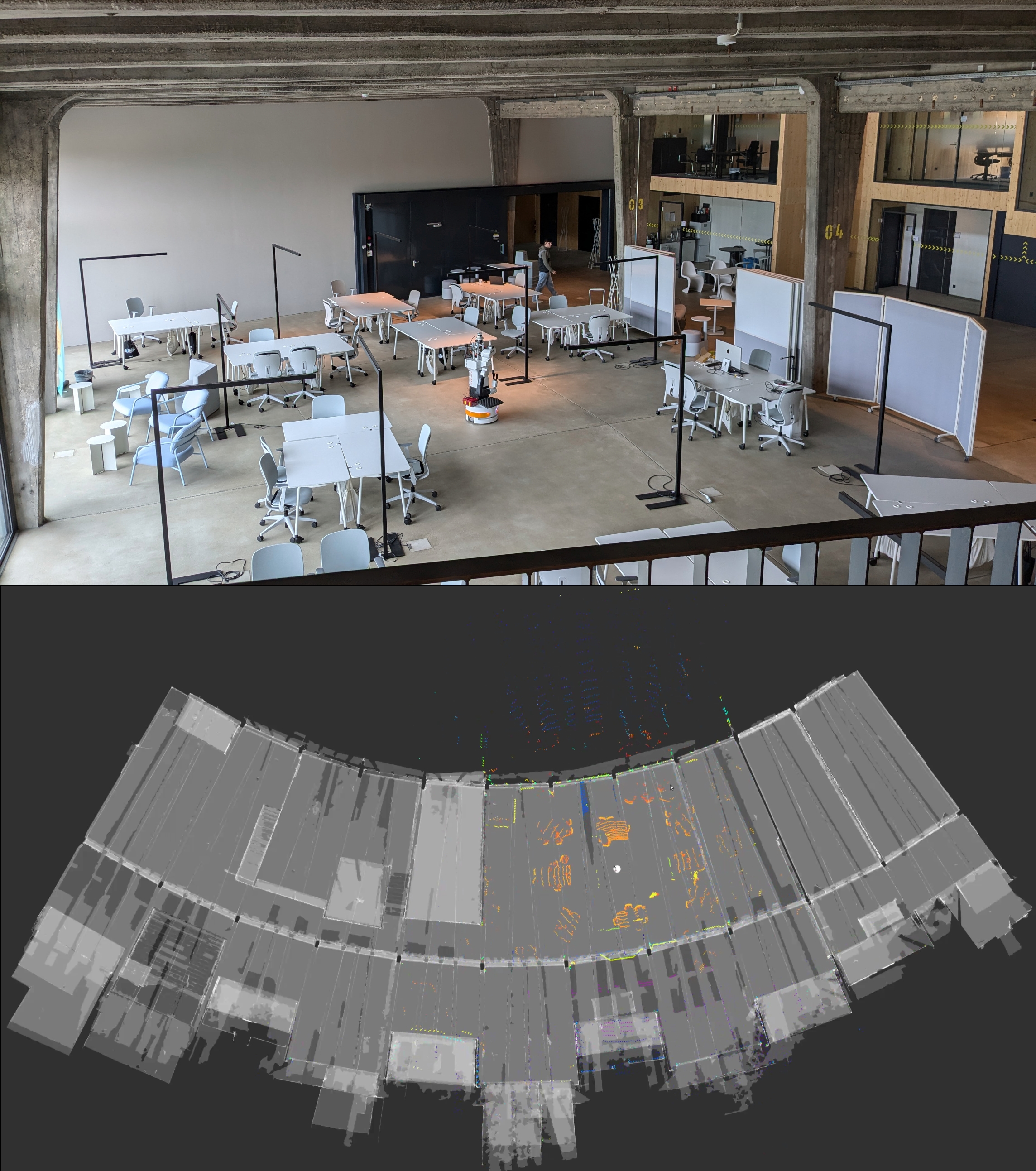}
    \caption{Localization of the TIAGo robot in a large-scale environment.
             The bottom image shows the pre-recorded polygonal map of the building (excluding furniture) overlaid with LiDAR points that were not associated with the map geometry during registration.}
    \label{fig:loc}
\end{figure}

Since our focus is on the development of spatio-semantic environment representations rather than a full SLAM system, we integrated the MICP-L approach~\cite{mock2024micp} on the TIAGo\footnote{\url{https://pal-robotics.com/es/robot/tiago/}} platform.
This enables us to utilize high-resolution pre-recorded maps (\eg from terrestrial LiDARs or CAD models) to obtain ground-truth-like pose estimations without requiring an external tracking system or running full SLAM.
This allows for the controlled transfer of our approach from simulated into real-world environments and enables the evaluation of the system with respect to noisy sensor and perception data.
The robot is equipped with an Ouster OS0 LiDAR for localization and a Femto Bolt ToF RGB-D camera.
Extrinsic calibration was performed by aligning the camera's point cloud to the 3D LiDAR reference frame via ICP registration.

Additionally, associating measurements with the known map geometry allows for the effective filtering of common panoptic classes (\eg walls, ceilings) that do not necessarily have to be tracked as separate instances in the semantic map.
This prunes unnecessary instances from the map and reduces uncertainty, thereby increasing overall efficiency (see Fig.~\ref{fig:loc}).

The robotic platform will be used to evaluate the full mapping and exploration pipeline, including \gls{3dssg} construction, view quality estimation, semantic querying, and spatial reasoning in future work.

\section{Discussion \& Outlook}
\label{sec:discussion}

The development of \acrshort{lierex} yielded several critical insights regarding the integration of \glspl{vlfm} into 3D exploration.

We observed that generating \emph{good} observation poses is rarely a function of geometry alone.
Simple geometric heuristics (\eg viewing angle or distance to centroid) fail in cluttered scenes where occlusion and semantic orientation (\eg the front of an object) are crucial.
Consequently, 2D map representations are insufficient; dense 3D information is essential to evaluate visibility and semantic relevance effectively.

Additionally, our initial experiments revealed that directly regressing optimal view poses, \eg via differentiable rendering, is impractical.
The non-convex nature of the optimization landscape frequently leads to convergence in local minima where views are geometrically valid but semantically meaningless.
This necessitated a pivot to a \emph{View Quality Estimation} approach, where the system regresses quality scores for sampled candidates rather than optimizing pose parameters directly.

Finally, we note a significant tension between instance-based and unified map representations.
While our \gls{3dssg} facilitates complex symbolic reasoning and open-set queries, maintaining geometric consistency for individual instances under sensor noise is significantly harder than in unified voxel-based representations (\eg TSDFs).
While unified maps offer superior efficiency for ray-casting and visibility checks, they often lack the flexibility required for the high-level semantic manipulation afforded by the graph structure.

In future work, we aim to address these trade-offs by integrating the learned view quality estimation into a hybrid, semantically guided frontier-based exploration planner.
This complete pipeline will be evaluated on the TIAGo demonstrator to validate the efficacy of open-set \gls{3dssg} reasoning in real-world scenarios.


\backmatter


\bmhead{Acknowledgements}

This work has been partially supported by the LIEREx project through a grant from the German Federal Ministry of Research, Technology and Space (BMFTR) with Grant Number 16IW24004.
The DFKI Niedersachsen (DFKI NI) is supported by the Ministry of Science and Culture of Lower Saxony and the Volkswagen Stiftung.

\section*{Declarations}


The authors have no competing interests to declare that are relevant to the content of this article.

\bibliography{bibliography}

\begin{thebibliography}{10}
\providecommand{\doi}[1]{\url{https://doi.org/#1}}
\bibcommenthead

\bibitem[\protect\citeauthoryear{N{\"u}chter and
  Hertzberg}{2008}]{nuchterSemanticMapsMobile2008b}
N{\"u}chter A, Hertzberg J.
\newblock Towards Semantic Maps for Mobile Robots.
\newblock Robotics and Autonomous Systems. 2008 Nov;56(11):915--926.
\newblock \doi{10.1016/j.robot.2008.08.001}.

\bibitem[\protect\citeauthoryear{Radford et~al.}{2021}]{radford2021learning}
Radford A, Kim JW, Hallacy C, Ramesh A, Goh G, Agarwal S, et~al.
\newblock Learning Transferable Visual Models from Natural Language
  Supervision.
\newblock In: International Conference on Machine Learning (ICML). PMLR; 2021.
  p. 8748--8763.

\bibitem[\protect\citeauthoryear{Rosinol et~al.}{2020}]{rosinol20203d}
Rosinol A, Gupta A, Abate M, Shi J, Carlone L.
\newblock {3D} Dynamic Scene Graphs: Actionable Spatial Perception with Places,
  Objects, and Humans.
\newblock Robotics: Science and Systems (RSS). 2020;.

\bibitem[\protect\citeauthoryear{Hughes et~al.}{2022}]{hughes2022hydra}
Hughes N, Chang Y, Carlone L.
\newblock {Hydra}: A Real-time Spatial Perception System for {3D} Scene Graph
  Construction and Optimization.
\newblock In: Robotics: Science and Systems (RSS); 2022. p. Article 50.

\bibitem[\protect\citeauthoryear{Liu et~al.}{2024}]{liu2024grounding}
Liu S, Zeng Z, Ren T, Li F, Zhang H, Yang J, et~al.
\newblock Grounding {DINO}: Marrying {DINO} with Grounded Pre-Training for
  Open-Set Object Detection.
\newblock In: European Conference on Computer Vision (ECCV). Springer; 2024. p.
  38--55.

\bibitem[\protect\citeauthoryear{Ren et~al.}{2024}]{ren2024grounding}
Ren T, Jiang Q, Liu S, Zeng Z, Liu W, Gao H, et~al.
\newblock Grounding {DINO} 1.5: Advance the "Edge" of Open-Set Object
  Detection.
\newblock arXiv preprint arXiv:240510300. 2024;\doi{10.48550/ARXIV.2405.10300}.
\newblock {\href{https://arxiv.org/abs/2405.10300}{{2405.10300}}}.

\bibitem[\protect\citeauthoryear{Zeng et~al.}{2023}]{zeng2023large}
Zeng F, Gan W, Wang Y, Liu N, Yu PS.
\newblock Large Language Models for Robotics: {A} Survey.
\newblock arXiv preprint arXiv:231107226. 2023;\doi{10.48550/ARXIV.2311.07226}.
\newblock {\href{https://arxiv.org/abs/2311.07226}{{2311.07226}}}.

\bibitem[\protect\citeauthoryear{Huang et~al.}{2024}]{huang2024large}
Huang D, Yan C, Li Q, Peng X.
\newblock From Large Language Models to Large Multimodal Models: A Literature
  Review.
\newblock Applied Sciences. 2024;14(12):5068.

\bibitem[\protect\citeauthoryear{Zhai et~al.}{2023}]{zhai2023sigmoid}
Zhai X, Mustafa B, Kolesnikov A, Beyer L.
\newblock Sigmoid Loss for Language Image Pre-Training.
\newblock In: Proceedings of the IEEE/CVF International Conference on Computer
  Vision (ICCV); 2023. p. 11975--11986.

\bibitem[\protect\citeauthoryear{Li et~al.}{2022}]{li2022blip}
Li J, Li D, Xiong C, Hoi S.
\newblock {BLIP}: Bootstrapping Language-Image Pre-Training for Unified
  Vision-Language Understanding and Generation.
\newblock In: International Conference on Machine Learning (ICML). PMLR; 2022.
  p. 12888--12900.

\bibitem[\protect\citeauthoryear{Li et~al.}{2023}]{li2023blip}
Li J, Li D, Savarese S, Hoi S.
\newblock {BLIP-2}: Bootstrapping Language-Image Pre-Training with Frozen Image
  Encoders and Large Language Models.
\newblock In: International Conference on Machine Learning (ICML). PMLR; 2023.
  p. 19730--19742.

\bibitem[\protect\citeauthoryear{Zhou et~al.}{2022}]{zhou2022extract}
Zhou C, Loy CC, Dai B.
\newblock Extract Free Dense Labels from {CLIP}.
\newblock In: European Conference on Computer Vision (ECCV). Springer; 2022. p.
  696--712.

\bibitem[\protect\citeauthoryear{Ghiasi et~al.}{2022}]{ghiasi2022scaling}
Ghiasi G, Gu X, Cui Y, Lin TY.
\newblock Scaling Open-Vocabulary Image Segmentation with Image-Level Labels.
\newblock In: European Conference on Computer Vision (ECCV). Springer; 2022. p.
  540--557.

\bibitem[\protect\citeauthoryear{Ding et~al.}{2023}]{ding2022open}
Ding Z, Wang J, Tu Z.
\newblock Open-Vocabulary Universal Image Segmentation with {MaskCLIP}.
\newblock In: International Conference on Machine Learning (ICML). vol. 202.
  {PMLR}; 2023. p. 8090--8102.

\bibitem[\protect\citeauthoryear{L{\"{u}}ddecke and
  Ecker}{2022}]{luddecke2022image}
L{\"{u}}ddecke T, Ecker AS.
\newblock Image Segmentation Using Text and Image Prompts.
\newblock In: IEEE/CVF Conference on Computer Vision and Pattern Recognition
  (CVPR). {IEEE}; 2022. p. 7076--7086.

\bibitem[\protect\citeauthoryear{Kirillov et~al.}{2023}]{kirillov2023segment}
Kirillov A, Mintun E, Ravi N, Mao H, Rolland C, Gustafson L, et~al.
\newblock {Segment Anything}.
\newblock In: Proceedings of the IEEE/CVF International Conference on Computer
  Vision (ICCV); 2023. p. 4015--4026.

\bibitem[\protect\citeauthoryear{Yamazaki et~al.}{2024}]{yamazaki2024open}
Yamazaki K, Hanyu T, Vo K, Pham T, Tran M, Doretto G, et~al.
\newblock {Open-Fusion}: Real-Time Open-Vocabulary {3D} Mapping and Queryable
  Scene Representation.
\newblock In: 2024 IEEE International Conference on Robotics and Automation
  (ICRA). IEEE; 2024. p. 9411--9417.

\bibitem[\protect\citeauthoryear{Maggio et~al.}{2024}]{maggio2024clio}
Maggio D, Chang Y, Hughes N, Trang M, Griffith D, Dougherty C, et~al.
\newblock {Clio}: Real-Time Task-Driven Open-Set {3D} Scene Graphs.
\newblock {IEEE} Robotics Autom Lett. 2024;9(10):8921--8928.
\newblock \doi{10.1109/LRA.2024.3451395}.

\bibitem[\protect\citeauthoryear{Laina et~al.}{2025}]{laina2025findanything}
Laina SB, Boche S, Papatheodorou S, Schaefer S, Jung J, Leutenegger S.
\newblock {FindAnything}: Open-Vocabulary and Object-Centric Mapping for Robot
  Exploration in Any Environment.
\newblock arXiv preprint arXiv:250408603. 2025;.

\bibitem[\protect\citeauthoryear{Kassab et~al.}{2024}]{kassab2024bare}
Kassab C, Mattamala M, Morin S, B{\"u}chner M, Valada A, Paull L, et~al.
\newblock The Bare Necessities: Designing Simple, Effective Open-Vocabulary
  Scene Graphs.
\newblock arXiv preprint arXiv:241201539. 2024;.

\bibitem[\protect\citeauthoryear{Linok et~al.}{2025}]{linok2025beyond}
Linok S, Zemskova T, Ladanova S, Titkov R, Yudin D, Monastyrny M, et~al.
\newblock Beyond Bare Queries: Open-Vocabulary Object Grounding with {3D} Scene
  Graph.
\newblock In: IEEE International Conference on Robotics and Automation (ICRA).
  IEEE; 2025. p. 13582--13589.

\bibitem[\protect\citeauthoryear{Zhao et~al.}{2023}]{zhao2023fast}
Zhao X, Ding W, An Y, Du Y, Yu T, Li M, et~al.
\newblock Fast {Segment Anything}.
\newblock arXiv preprint arXiv:230612156. 2023;.

\bibitem[\protect\citeauthoryear{Zhang et~al.}{2023}]{zhang2023faster}
Zhang C, Han D, Qiao Y, Kim JU, Bae SH, Lee S, et~al.
\newblock Faster {Segment Anything}: Towards Lightweight {SAM} for Mobile
  Applications.
\newblock arXiv preprint arXiv:230614289. 2023;.

\bibitem[\protect\citeauthoryear{Gadre et~al.}{2023}]{gadre2023cows}
Gadre SY, Wortsman M, Ilharco G, Schmidt L, Song S.
\newblock Cows on Pasture: Baselines and Benchmarks for Language-Driven
  Zero-Shot Object Navigation.
\newblock In: Proceedings of the IEEE/CVF Conference on Computer Vision and
  Pattern Recognition (CVPR); 2023. p. 23171--23181.

\bibitem[\protect\citeauthoryear{Chang et~al.}{2024}]{chang2024goat}
Chang M, Gervet T, Khanna M, Yenamandra S, Shah D, Min SY, et~al.
\newblock {GOAT}: {GO} to {Any Thing}.
\newblock In: Proceedings of Robotics: Science and Systems (RSS). Delft,
  Netherlands; 2024. p. Article 73.

\bibitem[\protect\citeauthoryear{Ren et~al.}{2024}]{ren2024explore}
Ren AZ, Clark J, Dixit A, Itkina M, Majumdar A, Sadigh D.
\newblock Explore until Confident: Efficient Exploration for Embodied Question
  Answering.
\newblock In: Proceedings of Robotics: Science and Systems (RSS). Delft,
  Netherlands; 2024. p. Article 89.

\bibitem[\protect\citeauthoryear{Gedicke et~al.}{2016}]{gedicke2016flap}
Gedicke T, G{\"u}nther M, Hertzberg J.
\newblock {FLAP} for {CAOS}: Forward-Looking Active Perception for
  Clutter-Aware Object Search.
\newblock IFAC-PapersOnLine. 2016;49(15):114--119.

\bibitem[\protect\citeauthoryear{Liu et~al.}{2024}]{liu2024okrobot}
Liu P, Orru Y, Vakil J, Paxton C, Shafiullah NMM, Pinto L.
\newblock Demonstrating {OK-Robot}: What Really Matters in Integrating
  Open-Knowledge Models for Robotics.
\newblock In: Proceedings of Robotics: Science and Systems (RSS). Delft,
  Netherlands; 2024. p. Article 91.

\bibitem[\protect\citeauthoryear{Wang et~al.}{2021}]{wang2021graspness}
Wang C, Fang HS, Gou M, Fang H, Gao J, Lu C.
\newblock {Graspness} Discovery in Clutters for Fast and Accurate Grasp
  Detection.
\newblock In: Proceedings of the IEEE/CVF International Conference on Computer
  Vision (ICCV); 2021. p. 15964--15973.

\bibitem[\protect\citeauthoryear{Fang et~al.}{2023}]{fang2023anygrasp}
Fang HS, Wang C, Fang H, Gou M, Liu J, Yan H, et~al.
\newblock {AnyGrasp}: Robust and Efficient Grasp Perception in Spatial and
  Temporal Domains.
\newblock IEEE Transactions on Robotics. 2023;39(5):3929--3945.

\bibitem[\protect\citeauthoryear{Savva et~al.}{2019}]{savva2019habitat}
Savva M, Kadian A, Maksymets O, Zhao Y, Wijmans E, Jain B, et~al.
\newblock {Habitat}: A Platform for Embodied {AI} Research.
\newblock In: Proceedings of the IEEE/CVF International Conference on Computer
  Vision (ICCV); 2019. p. 9339--9347.

\bibitem[\protect\citeauthoryear{Chang et~al.}{2017}]{chang2017matterport3d}
Chang A, Dai A, Funkhouser T, Halber M, Niessner M, Savva M, et~al.
\newblock {Matterport3D}: Learning from {RGB-D} Data in Indoor Environments.
\newblock arXiv preprint arXiv:170906158. 2017;.

\bibitem[\protect\citeauthoryear{Ramakrishnan
  et~al.}{2021}]{ramakrishnan2021habitat}
Ramakrishnan SK, Gokaslan A, Wijmans E, Maksymets O, Clegg A, Turner J, et~al.
\newblock {Habitat-Matterport 3D} Dataset ({HM3D}): 1000 Large-Scale {3D}
  Environments for Embodied {AI}.
\newblock arXiv preprint arXiv:210908238. 2021;.

\bibitem[\protect\citeauthoryear{Mock et~al.}{2024}]{mock2024micp}
Mock A, Wiemann T, P{\"u}tz S, Hertzberg J.
\newblock {MICP-L}: Mesh-Based {ICP} for Robot Localization Using
  Hardware-Accelerated Ray Casting.
\newblock In: 2024 IEEE/RSJ International Conference on Intelligent Robots and
  Systems (IROS). IEEE; 2024. p. 10664--10671.

\end{thebibliography}

\end{document}